# Small-world networks for summarization of biomedical articles

Milad Moradi

Section for Artificial Intelligence and Decision Support, Medical University of Vienna, Austria

Email: milad.moradivastegani@meduniwien.ac.at

**Abstract**

In recent years, many methods have been developed to identify important portions of text documents. Summarization tools can utilize these methods to extract summaries from large volumes of textual information. However, to identify concepts representing central ideas within a text document and to extract the most informative sentences that best convey those concepts still remain two crucial tasks in summarization methods. In this paper, we introduce a graph-based method to address these two challenges in the context of biomedical text summarization. We show that how a summarizer can discover meaningful concepts within a biomedical text document using the Helmholtz principle. The summarizer considers the meaningful concepts as the main topics and constructs a graph based on the topics that the sentences share. The summarizer can produce an informative summary by extracting those sentences having higher values of the degree. We assess the performance of our method for summarization of biomedical articles using the Recall-Oriented Understudy for Gisting Evaluation (ROUGE) toolkit. The results show that the degree can be a useful centrality measure to identify important sentences in this type of graph-based modelling. Our method can improve the performance of biomedical text summarization compared to some state-of-the-art and publicly available summarizers. Combining a concept-based modelling strategy and a graph-based approach to sentence extraction, our summarizer can produce summaries with the highest scores of informativeness among the comparison methods. This research work can be regarded as a start point to the study of small-world networks in summarization of biomedical texts.

**Keywords:** Biomedical text mining, Graph algorithms, Text summarization, Meaningfulness

## 1. Introduction

Managing and extracting useful information are demanding tasks that clinicians and biomedical researchers deal with when they work with large volumes of text documents. Automatic text summarization tools can reduce the time and effort needed to read and manage numerous and lengthy documents, specifically in the biomedical domain in which textual information is available by various resources and in different formats [1]. So far, different approaches have been adopted to the problem of text summarization. The graph-based approach has been studied in both domain-independent and biomedical text summarization research [2-4]. However, there are two non-trivial challenges that researchers have been trying to address: (1) discovering the main ideas and topics within the input text, and (2) extracting the most informative sentences to produce a summary that covers the main topics. In this paper, we address this two challenges in the context of biomedical text summarization.

We firstly extract biomedical concepts as building blocks of the input document, then identify the most important concepts using a meaningfulness measure and consider them as the main topics. Next, we construct a small-world network that conveys how the sentences are related in terms of the main topics. The reason for considering such type of text modeling is that having a graph with a small-world topology, it is possible to measure the contribution of nodes, i.e. sentences in our context, to the graph and make a reliable selection of the most important nodes [5]. We extract the sentences corresponding to the most important nodes to produce the final summary. Our method simply uses the degree to assess the importance of nodes and rank them. The results of evaluations show that our graph-based summarizer outperforms some comparison methods. Our graph-based approach can be a start point to the investigation of small-world networks, and related concepts and methods, for summarization of biomedical text documents.

The remainder of the paper is organized as follows. In Section 2, we have a brief overview on related work. Then we describe our graph-based summarization method in Section 3. The experimental results are presented and discussed in Section 4. Finally, we conclude and outline some future work in Section 5.

## 2. Related work

In recent years, many domain-independent summarizers have been proposed using different approaches [2]. Various methods employ generic features such as sentence length, sentence position, word frequency, presence of cue phrases, etc. to score the sentences of an input text and select the top ranked ones for the summary. It has been shown that these generic features may not be as useful as domain-specific methods in summarization of biomedical texts [1, 6-9]. Biomedical text documents have their own singularities [7] that require developing summarizers that make use of sources of domain knowledge. Our method utilizes concepts of the Unified Medical Language System (UMLS) [10] to analyze the input text in a concept-based level rather than considering only the terms. This can help to build a more accurate model from the ideas and topics within the text [1]. We show that, compared to domain-independent methods that use generic features, our domain-specific summarizer can achieve a higher performance in biomedical text summarization.

Biomedical summarizers have adopted different approaches from machine learning, statistics, Natural Language Processing (NLP), etc. to the problems of identifying the main topics and selecting the most relevant sentences [3]. These methods have been specialized for summarization of different types of documents including scientific papers, clinical notes, Electronic Health Records (EHRs), and so on [6]. The itemset-based summarizer [1] is a method evaluated for summarization of biomedical articles. It employs a data mining technique, i.e. frequent itemset mining, to discover the main topics within a text. It also uses a scoring formula to rank and select the most relevant sentences. The results of our evaluations show that our method can perform slightly better than the itemset-based summarizer.

In graph-based text summarization, different units of the text such as words, concepts, or sentences are considered as nodes and different types of relations are defined to draws edges [2, 5, 7]. The resulted graph represents the structure in which the units of the text are related together. Different scoring and ranking strategies are used to select the most important parts of the text based on the graph, its topology and properties, and the task at hand. Balinsky et al. [5] modeled text documents as small-world networks and showed that this type of modeling can be used in summarization. They did not conduct any evaluations that assess the performance of their approach against other methods for summarizing a corpus of texts. The current study is the first one that utilizes small-world networks for biomedical text summarization, also the first that evaluates this type of graph-based approaches by comparing against other methods on a corpus of documents.

## 3. Summarization method

Our summarization method consists of four main steps, i.e. (1) preprocessing, (2) finding meaningful concepts, (3) constructing the graph, and (4) summary generation. Fig. 1 illustrates the architecture of our graph-based biomedical text summarization method. We give a detailed description of each step in the following subsections.

### 3.1. Preprocessing

The preprocessing step begins by extracting the main body of the input document. This task is done according to the formatting and logical structure of the document. In our case, i.e. scientific articles, the main body is extracted by removing those parts of the text that seem to be unnecessary for inclusion in the summary. These parts may include the title, authors' information, abstract, keywords, header of sections and subsections, bibliography, and so on. The main body is splitted into several sentences represented by the set $\{S_1, S_2, \ldots, S_N\}$.

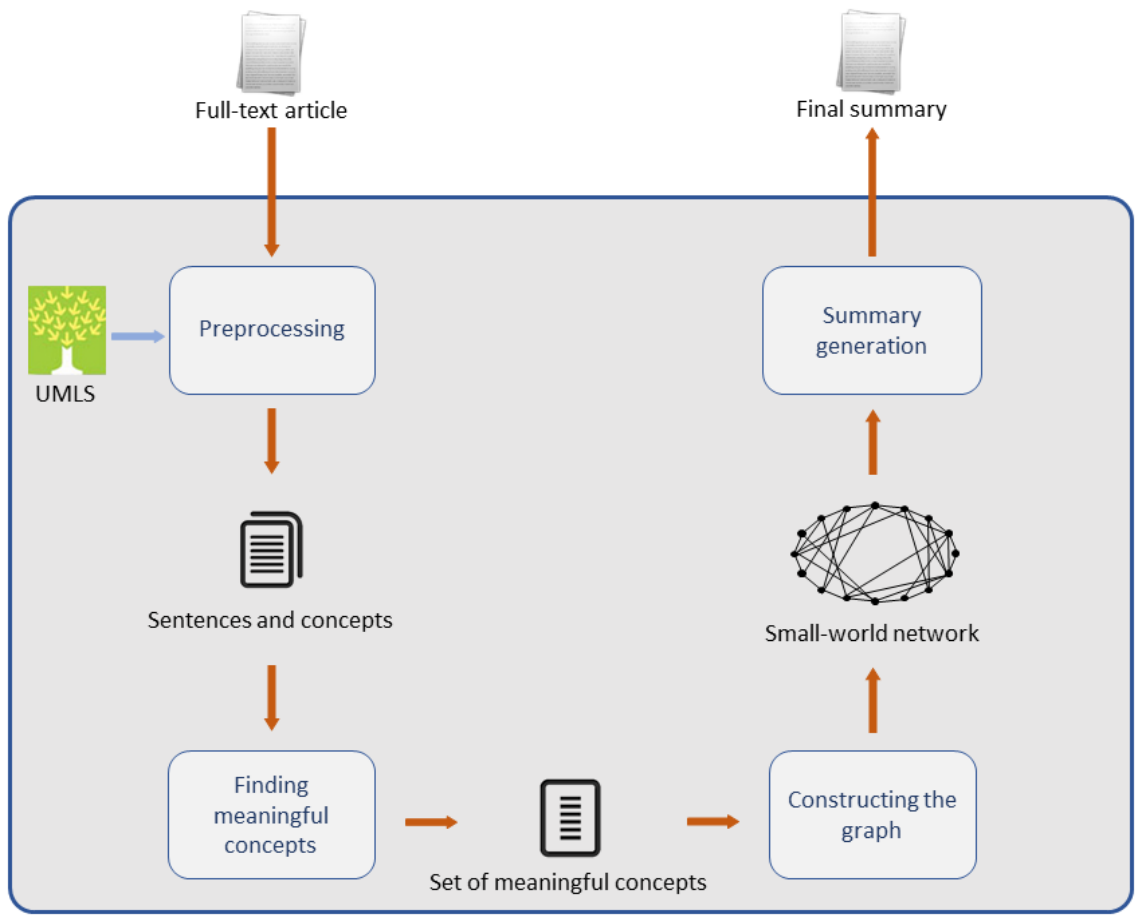

Fig. 1. The architecture of our graph-based biomedical text summarization method.

Afterwards, the text is mapped to the UMLS concepts using the MetaMap [11] tool that uses NLP methods and the UMLS resources [10] to identify noun phrases within each sentence and returns the best matching concepts. Every concept is associated with a semantic type that puts the concept into a broader semantic categorization. In fact, the semantic type determines the context in which the concept is more likely to appear in a specific phrase within the text. Fig. 2 shows a sample sentence and the UMLS concepts extracted in this step. At last, those concepts belonging to generic semantic types are discarded since they can be considered too broad and may not be useful in the analysis of a biomedical text [7]. These semantic types are *Temporal Concept*, *Spatial Concept*, *Qualitative Concept*, *Quantitative Concept*, *Language*, *Mental Process*, *Intellectual Product*, *Idea or Concept*, and *Functional Concept*. After the preprocessing step, each sentence $S_i$ is represented as a set of unique concepts.

### 3.2. Finding meaningful concepts

The Helmholtz principle from the Gestalt theory of human perception introduces a measure of meaningfulness that can be effectively used to discover rapid changes and unusual behavior in unstructured and text data [12]. In this step, we use the meaningfulness measure to identify the concepts stating main ideas within the text. Regarding the definition of the Helmholtz principle in the context of data mining, these concepts can be considered essential because they are observed in large deviations from randomness [6, 13].

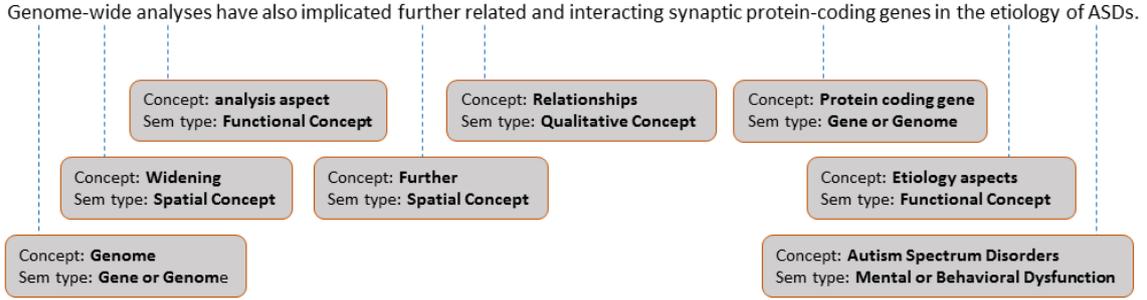

Fig. 2. A sample sentence and the UMLS concepts extracted in the preprocessing step.

Let $D$ be the input document and $P$ be a paragraph in $D$. For each concept $C_i$ in the concept set {$C_1$, $C_2$, ..., $C_M$} that contains all the unique concepts within $D$, we start to compute a meaningfulness value by calculating the Number of False Alarms (NFA) in every paragraph $P_j$. If concept $C_i$ appears $m$ times in $P_j$ and $k$ times in the whole document $D$, the NFA is computed as follows:

$$NFA(C_i, P_j, D) = \binom{k}{m} \frac{1}{N^{m-1}} \quad (1)$$

where $N$ is equal to [$L/B$] that $L$ is the total number of concepts within the document $D$, and $B$ is the total number of concepts in the paragraph $P_j$. In Eq. 1, $\binom{k}{m}$ is a binomial coefficient computed as follows:

$$\binom{k}{m} = \frac{k!}{m!\,(k-m)!} \quad (2)$$

Afterwards, the meaningfulness value of the concept $C_i$ inside the paragraph $P_j$ is computed as follows:

$$Meaning(C_i, P_j, D) = -\frac{1}{m} \log NFA(C_i, P_j, D) \quad (3)$$

Eventually, we construct a set *MeaningfulSet(ε)* holding the meaningful concepts. The concept $C_i$ is added to the set *MeaningfulSet(ε)* if the value of $Meaning(C_i, D)$ is greater than $\varepsilon$. The value of $Meaning(C_i, D)$ is the maximum of values $Meaning(C_i, P, D)$ over all paragraphs within $D$, and $\varepsilon$ is a parameter that determines the level of meaningfulness. At the end of this step, we have a set of meaningful concepts being regarded as the main ideas within the input document.

As an example, Table 1 presents the meaningful concepts identified within a sample document[1]. The sample document is a scientific article related to the genetic overlap of three mental disorders. It contains 85 sentences. In this example, the value of the meaningfulness parameter $\varepsilon$ is set to 0.2. Note that this is an example, and the optimum value for the parameter $\varepsilon$ will be specified in Section 4.1.

---

[1] Available at https://genomemedicine.biomedcentral.com/articles/10.1186/gm102

**Table 1.** The meaningful concepts identified within the sample document for the meaningfulness level of 0.2. The concepts are presented in descending order of their meaning values.

| Concept | Meaning | Concept | Meaning |
|---|---|---|---|
| Industrial machine | 1.445 | Binding (Molecular Function) | 0.741 |
| Promotion (action) | 1.445 | Procedure findings | 0.711 |
| Inhibition | 1.319 | Genes | 0.587 |
| Gene Knockout Techniques | 1.319 | Reporting | 0.524 |
| ethnic european | 1.282 | Single Nucleotide Polymorphism | 0.520 |
| Neurodevelopmental disorder | 1.282 | Phenotype | 0.520 |
| Research Activities | 1.282 | neuroligin | 0.509 |
| Mental disorders | 1.262 | Mutation | 0.490 |
| genetic association | 1.242 | Copy Number Polymorphism | 0.411 |
| Synapses | 0.991 | SHANK3 gene | 0.265 |
| APBA2 gene | 0.867 | Overlap | 0.235 |
| synaptogenesis | 0.770 | | |

### 3.3. Constructing the graph

At this step, the input document is modeled as a graph $G = (V, E)$, where $V$ is the set of vertices, i.e. the sentences of the document $D$, and $E$ is the set of edges representing the relationships between the sentences. There are two types of edges in the graph $G = (V, E)$ that represent two types of relations, i.e. local and distant [5]. Local relations are modeled by the edges connecting every pair of consecutive sentences $S_i$ and $S_{i+1}$. The idea behind local relations is that the order in which sentences appear in a document can be important to model the document, as well as consecutive sentences are usually related. Distant relations are represented by the edges connecting two non-consecutive sentences that share some topics [5]. In our case, the topics are those meaningful concepts identified in the previous step. Therefore, there is an edge between two given vertices if the corresponding sentences have at least one concept from the set *MeaningfulSet(ε)* in common. The number of edges depends on the meaningfulness parameter $ε$. At this point, our graph $G = (V, E)$ represents the sentences and relations existing among them inferred with respect to the meaningful concepts.

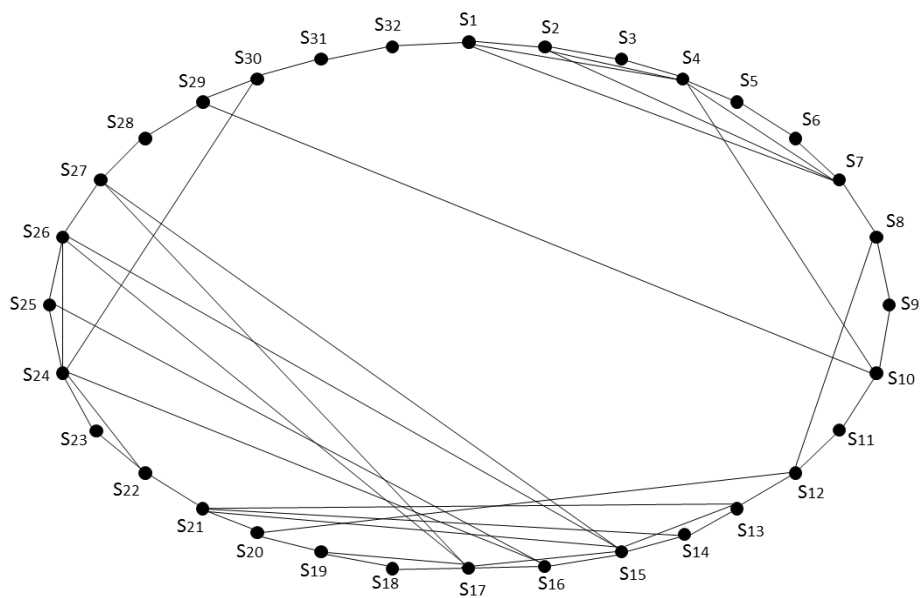

Fig. 3. The graph constructed for the first 32 sentences of the sample document.

As an example, Fig. 3 shows the graph constructed for the sample document. For clarity and brevity reasons, this example shows only the nodes corresponding to the first 32 sentences. The edges between non-consecutive nodes are drawn according to the meaningful concepts presented in Table 1.

### 3.4. Summary generation

After creating the graph, the summarizer needs a method to identify the most important sentences. The degree of each node can be used as a measure of centrality to assess its importance. The higher the degree associated with a node, the more the number of sentences having some main topics in common with the corresponding sentence. Therefore, those sentences with a higher degree can be considered more informative.

The summarizer computes the degree of each node and ranks them in descending order. It selects the top *N* nodes from the ranked list and extracts the corresponding sentences. *N* is the number of sentences that should be selected for the summary and is computed based on the compression rate. For example, if the compression rate is equal to 0.3, almost 26 sentences will be extracted from the sample document to generate a summary. The summarizer arranges the summary's sentences into the order in which they appear in the input document. As an example, Table 2 presents the top-ranked 26 sentences of the sample document. These sentences are extracted to produce the final summary.

## 4. Experimental results

We conduct two sets of experiments to assess the performance of our method for summarization of biomedical articles. In the first set, we evaluate the summarization method under different values for the meaningfulness parameter $\varepsilon$. The second set of experiments is devoted to the comparison of our method with other summarizers.

**Table 2.** The top-ranked 26 sentences of the sample document. The sentences are ranked based on the degree values for the corresponding nodes.

| Rank | Sentence | Degree | Rank | Sentence | Degree |
|------|----------|--------|------|----------|--------|
| 1    | $S_{15}$ | 22     | 14   | $S_{73}$ | 16     |
| 2    | $S_{38}$ | 20     | 15   | $S_{83}$ | 16     |
| 3    | $S_{24}$ | 19     | 16   | $S_{35}$ | 15     |
| 4    | $S_{33}$ | 19     | 17   | $S_{57}$ | 15     |
| 5    | $S_{54}$ | 19     | 18   | $S_{62}$ | 15     |
| 6    | $S_{55}$ | 19     | 19   | $S_{64}$ | 15     |
| 7    | $S_{53}$ | 18     | 20   | $S_{66}$ | 15     |
| 8    | $S_{16}$ | 17     | 21   | $S_{67}$ | 15     |
| 9    | $S_{61}$ | 17     | 22   | $S_{68}$ | 15     |
| 10   | $S_{70}$ | 17     | 23   | $S_{69}$ | 15     |
| 11   | $S_{25}$ | 16     | 24   | $S_{75}$ | 15     |
| 12   | $S_{48}$ | 16     | 25   | $S_{14}$ | 14     |
| 13   | $S_{63}$ | 16     | 26   | $S_{80}$ | 13     |

We use the Recall-Oriented Understudy for Gisting Evaluation (ROUGE) toolkit [14] in our experiments. The ROUGE compares a system-generated summary with a model summary and computes some scores conveying the content overlap. We use two ROUGE metrics, i.e. ROUGE-2 (R-2) and ROUGE-SU4 (R-SU4), in our evaluations. R-2 computes the number of shared bigrams and R-SU4 computes the overlap of skip-bigrams with a skip distance of four. For the evaluation of different methods, we randomly selected 300 biomedical articles from the BioMed Central's corpus for text mining research [15]. Each summarizer produces a summary for every article in the corpus, and we use the abstracts of articles as the model summaries. In order to specify the value of the meaningfulness parameter, we use a separate development corpus of 100 articles. The compression rate in all the experiments is equal to 0.3. We use a Wilcoxon signed-rank test with a confidence interval of 95% to test the statistical significance of results.

### 4.1. The meaningfulness parameter

As mentioned in Section 3.2, there is a parameter $\varepsilon$ used by the summarization method to identify the meaningful concepts within the input text. When constructing the graph, if two sentences have at least a meaningful concept in common, the method draws an edge between the corresponding nodes. Therefore, if the value of the parameter $\varepsilon$ is too small, we will have a large number of meaningful concepts and the network will be a large random graph with numerous edges. On the other hand, if the value of the parameter $\varepsilon$ is too large, we will have a small number of meaningful concepts and the network will be a regular graph that has only local relations. However, there is a range of values of the parameter $\varepsilon$ that the graph will be a small-world network. In this range, the expected behavior of our small-world network is defined by four properties [5]: (1) relatively small number of edges, (2) small degree of separation, (3) high mean clustering, and (4) high transitivity.

In order to find out the range of the meaningfulness parameter $\varepsilon$ in which the graph has the behavior of a small-world network, we assess the above four properties for the graphs resulted from the documents of the development corpus. We assign different values to the parameter $\varepsilon$ and assess these properties of the graphs: the number of edges, the characteristic path length, the clustering coefficient (mean clustering), and the transitivity. The graphs have the behavior of a small-world network when the value of $\varepsilon$ is in the range [0.1, 0.8]. Table 3 presents the R-2 and R-SU4 scores obtained by the graph-based summarizer when a value from the above range is assigned to the meaningfulness parameter. The difference between the scores is not usually significant ($p>0.05$) when we assign any value from the above range. Since the summarizer obtains the highest scores on the development corpus when the value of $\varepsilon$ is equal to 0.3, we use this as the optimal value in the final evaluations.

**Table 3.** ROUGE scores obtained by the graph-based summarizer using different values of the meaningfulness parameter. The best score for each ROUGE metric is shown in bold type.

| Meaningfulness parameter | ROUGE-2 | ROUGE-SU4 |
| --- | --- | --- |
| 0.1 | 0.3306 | 0.3753 |
| 0.2 | 0.3341 | 0.3779 |
| 0.3 | **0.3397** | **0.3802** |
| 0.4 | 0.3363 | 0.3784 |
| 0.5 | 0.3318 | 0.3741 |
| 0.6 | 0.3280 | 0.3705 |
| 0.7 | 0.3246 | 0.3689 |
| 0.8 | 0.3212 | 0.3654 |

### 4.2. Comparison with other summarizers

We compare the performance of our graph-based method with four summarizers. The itemset-based summarizer [1] is a method for summarization of biomedical texts. It extracts frequent itemsets from the concepts of the input document, then assigns a score to each sentence according to the presence of the itemsets. It selects sentences with the highest scores for inclusion in the summary. SUMMA [16] uses generic features such as term frequency and sentence position to score sentences and select them for the summary. MEAD [17] generates summaries based on centroid, positional, length, and term similarity features. TexLexAn [18] uses keywords extracted from the input text and a set of cue expressions to select the most relevant sentences and return the summary. We run the comparison methods using their best settings specified on the development corpus. Table 4 presents the ROUGE scores for our graph-based summarizer and other comparison methods.

**Table 4.** ROUGE scores obtained by our graph-based method and the other summarizers. The best score for each ROUGE metric is shown in bold type.

|  | **ROUGE-2** | **ROUGE-SU4** |
|---|---|---|
| Graph-based summarizer | **0.3321** | **0.3753** |
| Itemset-based summarizer | 0.3258 | 0.3697 |
| SUMMA | 0.2959 | 0.3485 |
| MEAD | 0.2847 | 0.3402 |
| TexLexAn | 0.2770 | 0.3296 |

As the results show, our graph-based method obtains the highest scores among the comparison methods. Compared to SUMMA, MEAD, and TexLexAn, our summarizer significantly improves both the R-2 and R-SU4 scores ($p<0.05$). This shows that our approach that uses the meaningfulness measure to construct a small-world network from the input text, and extracts the most informative sentences simply based on the degree, can be more useful than generic features, such as position, length, keyword, etc. for summarization of biomedical articles.

The scores obtained by the graph-based method are slightly better than those of the itemset-based summarizer, but the improvement is not significant for both the scores ($p>0.05$). This shows that the performance of the simple approach adopted by the graph-based summarizer can be comparable to some state-of-the-art methods in biomedical text summarization. The itemset-based summarizer discovers the main topics in the form of frequent itemsets containing important concepts and uses them to extract the most informative sentences. The graph-based method identifies the main ideas in the form of meaningful concepts, utilizes them to construct a small-world of sentences that conveys the relationships between sentences, and uses the degree to rank the sentences and select the most important ones. These two methods employ different approaches to the problems of identifying main topics within the input text and extracting the most informative sentences. However, both the summarizers perform well in summarization of biomedical articles regarding the results presented by Table 4.

## 5. Conclusion and future work

In this paper, we introduced a graph-based approach to biomedical text summarization. Our method mapped the input document to the UMLS concepts and used a meaningfulness measure to identify the most important concepts that convey main ideas and topics within the text. It constructed a small-world network based on the meaningful concepts that sentences share with each other. The summarizer used the degree measure to assess the importance of sentences and selected the top-ranked ones for the summary. The results showed that this relatively simple approach can be effectively employed to produce highly informative summaries for biomedical articles. This study can be extended by utilizing more complex methods in different steps of the summarization process. Other measures can be investigated for discovering main topics in a document, as addressed to some extent by previous research [6]. More advanced strategies for constructing the graph can be developed to have different structures that may model the input text more accurately. Moreover, various centrality measures and ranking methods can be evaluated and developed to have other metrics that may assess the significance of sentences more effectively than the simple degree.